\documentclass[10pt,twocolumn,letterpaper]{article}
\pdfoutput=1
\usepackage{wacv}
\usepackage{times}
\usepackage{epsfig}
\usepackage{graphicx}
\usepackage{amsmath}
\usepackage{amssymb}
\usepackage{url}

\wacvfinalcopy 

\def\wacvPaperID{47} 

\ifwacvfinal\fi
\begin{document}

\title{Detecting the Starting Frame of Actions in Video}

\author{Iljung S. Kwak\\
UC, San Diego\\
Janelia Research Campus\\
{\tt\small kwaki@janelia.hhmi.org}
\and
Jian-Zhong Guo\\
Janelia Research Campus\\
{\tt\small guoj@janelia.hhmi.org}
\and
Adam Hantman\\
Janelia Research Campus\\
{\tt\small hantmana@janelia.hhmi.org}
\and
David Kriegman\\
UC, San Diego\\
{\tt\small kriegman@cs.ucsd.edu}
\and
Kristin Branson\\
Janelia Research Campus\\
{\tt\small bransonk@janelia.hhmi.org}
}

\maketitle
\ifwacvfinal\thispagestyle{empty}\fi

\begin{abstract}
In this work, we address the problem of precisely localizing key frames of an action, for example, the precise time that a pitcher releases a baseball, or the precise time that a crowd begins to applaud. Key frame localization is a largely overlooked and important action-recognition problem, for example in the field of neuroscience, in which we would like to understand the neural activity that produces the start of a bout of an action. To address this problem, we introduce a novel structured loss function that properly weights the types of errors that matter in such applications: it more heavily penalizes extra and missed action start detections over small misalignments. Our structured loss is based on the best matching between predicted and labeled action starts. We train recurrent neural networks (RNNs) to minimize differentiable approximations of this loss. To evaluate these methods, we introduce the Mouse Reach Dataset, a large, annotated video dataset of mice performing a sequence of actions. The dataset was collected and labeled by experts for the purpose of neuroscience research. On this dataset, we demonstrate that our method outperforms related approaches and baseline methods using an unstructured loss.
\end{abstract}

\section{Introduction}
Video-based action recognition tasks are generally framed in one of two ways. In action classification \cite{karpathy2014large,soomro2012ucf101,idrees2017thumos}, the goal is to assign a single category to a trimmed video. In fine-grained action detection or segmentation~\cite{idrees2017thumos,Heilbron_2015_CVPR}, the goal is to assign a category to each frame in the video. In this work, we focus on the relatively unexplored problem of detecting and classifying key frames, such as the starts of action bouts. 

\begin{figure}[t]
\begin{center}
\includegraphics[height=0.90\linewidth]{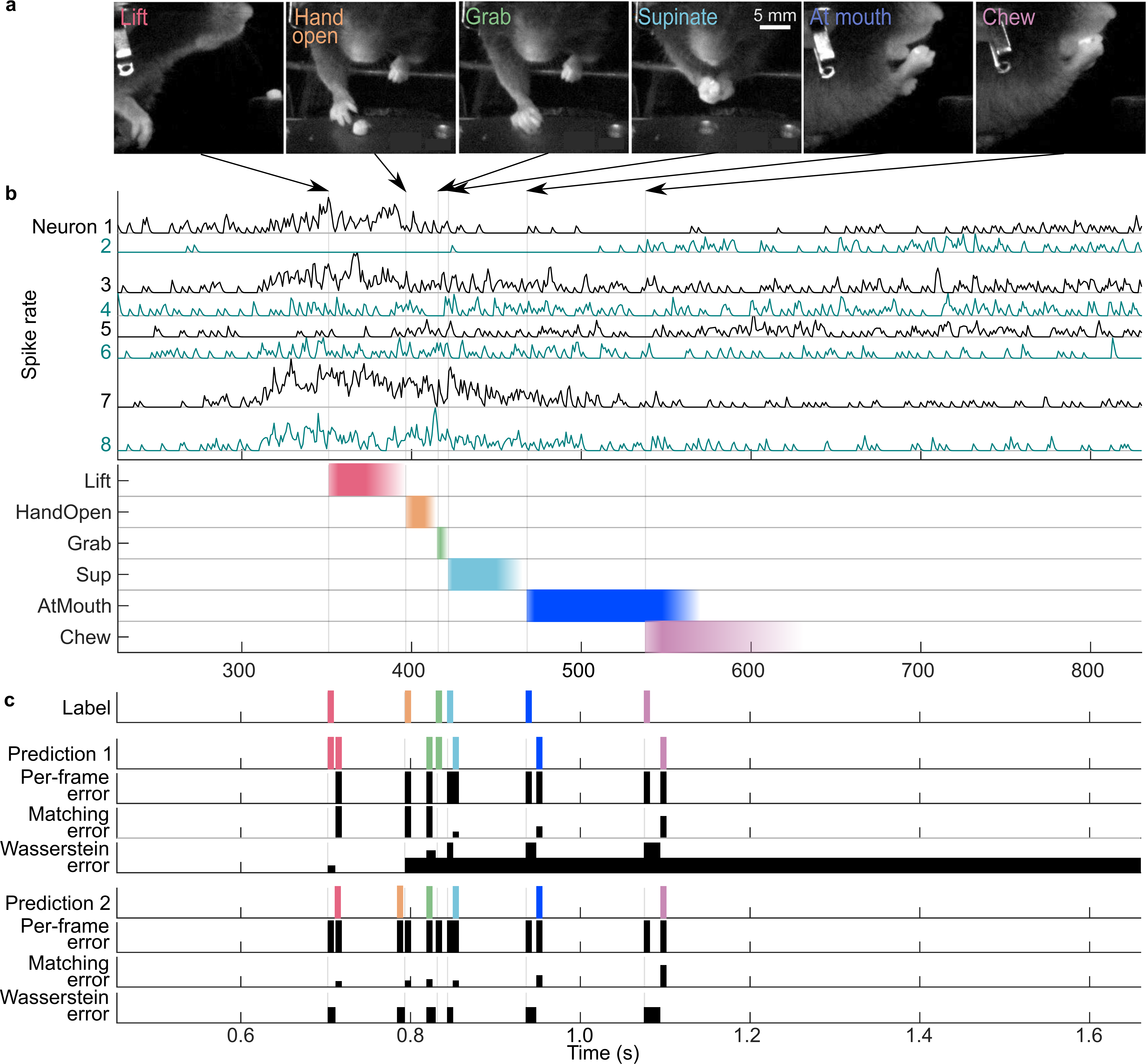}
\end{center}
   \caption{(a) In this neuroscience experiment, a mouse has been trained to reach for a food pellet. This movement consists of a sequence of actions: lift, hand-open, grab, supinate, pellet at-mouth, and chewing~\cite{guo2015cortex}. (b) A fundamental goal in systems neuroscience is to associate patterns of neural activity (top) with the behaviors it causes (bottom), e.g. spiking in several of the recorded cortical neurons precedes the onset of lift. Colors indicate different behaviors, and saturation indicates annotator confidence. Confidence changes are sharper at action starts than ends, as starts are usually associated with large accelerations, e.g. pinpointing the start of a lift is much easier than pinpointing its end. (c) Given a sequence of labeled frames (Labels), a per-frame loss prefers multiple or missed detections (Prediction 1) to a small temporal offset in the predictions (Prediction 2). The structured losses proposed in this work are designed to instead heavily penalize extra or missed detections. Error plots (black) show the error accrued on each frame.}
\label{fig:overview}
\vspace{-0.4cm}
\end{figure}

This is an important problem for many computer-vision applications, including neuroscience research. A fundamental goal in neuroscience is to understand the neural activity patterns that produce behavior. To do this, researchers localize the {\em starts} of bouts of actions, then examine neural activity just prior to this \cite{sauerbrei2018cortical} (Figure~\ref{fig:overview}a-b). In this work, we focus on detecting action starts, although the proposed methods could be applied to any identified key frame in an action bout. 

Videos often contain many bouts of the same action, which can happen one after the other, e.g. individual strides while walking, or multiple attempts to grab an object. In such applications, extra action start detections (false positives) or missing action start detections (false negatives) are much more costly than a predicted start offset from the true start by a small number of frames. Using an unstructured, per-frame error between the true and predicted action starts would incorrectly penalize a detection being offset by a small number of frames from the ground truth \textit{more} than having a false positive or negative (Figure~\ref{fig:overview}c). In this work, we propose a structured loss that involves finding the best match between action start predictions and labels, allowing us to properly weight each of these types of errors. We propose using a recurrent neural network (RNN) to minimize this structured loss using gradient descent. As this loss is not differentiable, we also propose to minimize a differentiable proxy based on the Earth Mover's Distance (EMD).

We introduce a new video data set, The Mouse Reach Dataset, that has been annotated with the starting frames of a set of behaviors. This data set consists of videos of mice performing a task that starts with reaching for a food pellet and ends with chewing that pellet; when successful, the task consists of a sequence of six actions (Figure~\ref{fig:overview}a). The sequence has strong temporal structure that can be exploited by an RNN, but can also vary substantially. Actions may be repeated, such as when a mouse fails to grab the food pellet on the first attempt and then tries again. These action categories were chosen by biologists, and their starts can be consistently identified. Furthermore, reaching tasks are often used in rodents and primates to study motor control, learning, and adaptation, and tools for automatic quantification of reach behavior would have immediate impact on neuroscience research~\cite{whishaw1990structure,guo2015cortex,sauerbrei2019cortical,krakauer2019motor}. 

We show that an RNN trained to minimize either of our structured losses outperforms an RNN trained to minimize the unstructured per-frame loss. Furthermore, our method outperforms a recently developed, competing algorithm for action start detection~\cite{shou2018online}. We also show that our proposed method outperforms baseline and competing methods on the THUMOS'14 data set~\cite{THUMOS14}.

In summary, we introduce a novel structured loss function and show how RNNs can be used to minimize this, and contribute a new, real-world dataset for fine-grained action start detection that has been annotated in the course of neuroscience research. We describe our algorithm in Sec.~\ref{sec:methods}, our dataset in Sec.~\ref{sec:dataset}, and experimental results in Sect.~\ref{sec:experiments}.

\section{Related Work}
\vspace{-.2cm}
\label{sec:relatedwork}
Our work has many similarities to fine-grained action detection, in which the goal is to categorize the action at each frame. We emphasize that our error criterion differs from these approaches. In fine-grained action detection, to incorporate the temporal context, 3D convolutional networks and recurrent networks have been used to detect actions~\cite{shou2017cdc,xu2017r,singh2016multi,wei2018sequence}. Following the success of region proposals~\cite{uijlings2013selective,ren2015faster} for object detection, algorithms for proposing temporal segments for action classification have been developed~\cite{gao2017turn,escorcia2016daps}. 

Fine-grained action detection algorithms have leveraged feature representations first developed for trimmed action recognition~\cite{wei2018sequence,escorcia2016daps,gao2017turn}. Large-scale action recognition datasets~\cite{carreira2017quo,karpathy2014large} have helped produce strong representations of short video snippets, which then can be used by detection algorithms. 3D convolutional networks leverage lessons learned from successful image recognition networks~\cite{tran2015learning} and simultaneously learn appearance and motion information. Recurrent models, such as LSTMs, have been used to model long range temporal relationships~\cite{karpathy2014large}. More recently, two stream networks~\cite{simonyan2014two,wang2016temporal,carreira2017quo} have been successful at action recognition. In our work we use two-stream feature representations as inputs to our detection model.

Online detection of action start (ODAS)~\cite{shou2018online} is the most similar past work to ours. In ODAS, the goal is to accurately detect the start of an action for use in real-time systems. In contrast, our work focuses on offline detection of action starts to understand the causes of behaviors, for example to understand the neural activity that produced a behavior. Both offline and online start detection have similar difficulties in label sparsity. In this work, we provide a dataset for which the accuracy of the action start labels was the main focus in dataset creation. Our dataset will be useful for both online and offline action start detection research.

\begin{figure}[htb]
\begin{center}
\includegraphics[width=0.95\linewidth]{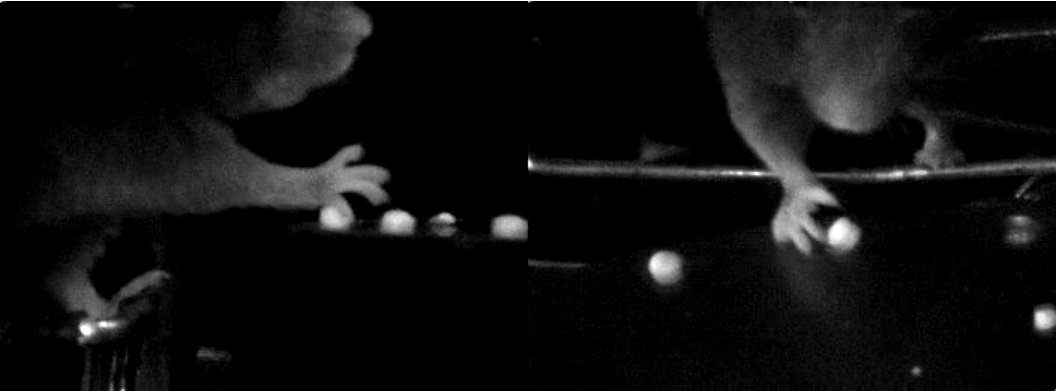}
\end{center}
   \caption{Our dataset contains two recorded views of the mouse, one from the side (left) and one from in front of the mouse (right). In this pair of frames, the mouse is attempting to grab a food pellet.}
\label{fig:dataset}
\end{figure}

\vspace{-0.3cm}
\section{Methods}
\label{sec:methods}
\subsection{Problem Formulation}
Let $\mathbf{X} = [x^{1}, x^{2}, ..., x^{T}]$ be a sequence of $T$ video frames, where $x^{t} \in \mathbf{R}^d$ is the feature representation of each frame. The goal of our work is to predict, for each frame $t$ and behavior $b$, whether the frame corresponds to the start of a behavior bout ($y^t_b = 1$) or not ($y^t_b = 0$). Let $\mathbf{Y} = [y^{1}, y^{2}, ..., y^{T}]$ be the sequence of ground truth labels for $X$, where $y^{t} \in \{0, 1\}^B$ and $B$ is the number of behaviors. 

Let $\hat{Y}=[\hat{y}^1, \hat{y}^2, ..., \hat{y}^T]$ be a predicted sequence of labels. We propose to measure the following structured error. We match behavior starts $y^i_b = 1$ with predictions $\hat{y}^{j}_b = 1$. Each label can be matched with at most one prediction within $|i-j| < \tau$ frames. Labeled starts without a matched predicted start are false negatives (FN) and get a fixed penalty of $\tau$. Predicted starts without a matched true start are false positives (FP) and get a fixed penalty $\tau$.

More formally, let $M\in\mathbf{Z}^{T\times B}$ be a matching from true to predicted starts, where $m^i_b > 0$ and $y^i_b=1$ means that the true start of behavior $b$ at frame $i$ is matched to a predicted start of behavior $b$ at frame $m^i_b$, and $m^i_b = 0$ indicates that the true start at frame $i$ is not matched. Similarly, let $\bar{M}$ denote an inverse matching from predictions to labels consistent with $M$. Then, our error criterion can be written as a minimum over matchings $M$:

\begin{equation}
\begin{array}{rl}
Err(Y,\hat{Y}) = & \\
\min_{M}\sum_{tb} & [ \underbrace{I(y^t_b=1) (I(m^t_b=0)\tau)}_{\text{FN}} + \\
          &\underbrace{I(m^t_b > 0)|t - m^t_b|}_{\text{TP}}) + \\
          &\underbrace{I(\hat{y}^t_b=1) I(\bar{m}^t_b=0)\tau)}_{\text{FP}}]
\label{eq:matchcost}
\end{array}
\end{equation}
where $I()$ is the indicator function.


We can use the Hungarian algorithm~\cite{kuhn55} to efficiently compute the optimal matching $M$ in this criterion. For each behavior, our bipartite graph consists of two sets of $N+M$ nodes, where $N$ and $M$ are the number of true and predicted action starts. In the first set, the first $N$ nodes correspond to true starts and the last $M$ nodes correspond to false positives. In the second set, the first $M$ nodes correspond to predicted starts and the last $N$ to false negatives. The distance matrix is then
$$ D^{nm} = \left\{ \begin{array}{rl}
|s^n-\hat{s}^m| & n \leq N, m \leq M \text{ (TP)}\\
\tau & n > N, m \leq M \text{ (FP)}\\
\tau & n \leq N, m > M\text{ (FN)}\\
\tau & n > N, m > M
\end{array} \right.,
$$
where $s^n$ is the $n$th true action start and $\hat{s}^m$ is the $m$th predicted action start.


\subsection{Matching Loss}
\label{sec:MatchingLoss}
We propose a structured loss based on this error criterion, which we refer to as the Matching Loss. The output of our classifiers are continuous values $\hat{y}^t_b \in [0,1]$. To compute this loss, we binarize the classifier outputs by thresholding and apply non-maximal suppression, resulting in a sequence of predicted action starts $\hat{S}$. We use $\hat{S}$ to select an optimal matching $\hat{M}$ using the Hungarian algorithm, as described above. Then, we minimize the following loss, which is a differentiable function of the continuous classifier outputs:
\begin{equation}
\begin{array}{rl}
\mathcal{L}_H(Y,\hat{Y},\hat{M}) = &\\
\sum_{tb} & C_b [ \underbrace{I(y^t_b=1) (I(\hat{m}^t_b=0) C_{fn}}_{\text{FN}} - \\
          & \underbrace{I(\hat{m}^t_b > 0)(\tau-|t - m^t_b|) \hat{y}^t_b C_{tp}}_{\text{TP}} +  \\
          & \underbrace{\hat{y}^t_b I(\bar{m}^t_b=0)C_{fp}}_{\text{FP}}]
\label{eq:matching}
\end{array}
\end{equation}
where $C_b$ is a weight for behavior $b$ (usually set to one over the number of true starts of that behavior). $C_{tp}$, $C_{fn}$, and $C_{fp}$ are parameters for weighing the importance of true positives, false negatives, and false positives respectively.

%

Note that our loss can be applied to any matching, but we choose to use the optimal matching. With this loss, we can directly enforce the importance of predicting near the true behavior start frame while avoiding spurious predictions. A correct prediction is penalized by the distance to the true behavior start frame and the confidence of the network output. Any prediction that is not matched is be penalized by the network's output score.

Given the matching $\hat{M}$, this loss is differentiable and thus can be minimized using gradient descent. Following \cite{stewart2016end, wei2018sequence}, we iteratively hold fixed the network and select the optimal matching, then fix the matching and apply gradient descent to optimize the network. In our experiments, we found that, using this training procedure, our networks were able to learn to localize behavior start location.However, selecting the optimal matching $\hat{M}$ is not differentiable.

\subsection{Wasserstein/EMD Loss}
\label{sec:WassersteinLoss}
Our Matching Loss relies on an assignment of predictions to ground truth labels. The assignment problem can be formulated probabilistically and solved with the Wasserstein Distance~\cite{peyre2019computational}. Similar to \cite{wei2018sequence}, we use the squared EMD loss, a variant of the 1-Wasserstein Distance, as an alternative structured cost for the sequence. Unlike \cite{wei2018sequence}, we do not apply a matching before computing the loss. This allows the loss to be completely differentiable. Additionally, the Wassertein loss is applied to all predictions for a behavior simultaneously, rather than to each prediction separately. The predicted label sequence and the ground truth label sequence are first normalized:
\begin{equation}
    \begin{array}{ll}
        y_b'^i &= \frac{1}{\sum_{i=0}^T (y_b^i + \epsilon)}(y_b^i + \epsilon)\\
        \hat{y}_b'^i &= \frac{1}{\sum_{i=0}^T (\hat{y}_b^i + \epsilon)}\hat{y}_b^i\\
    \end{array}.
\label{eq:wasser}
\end{equation}
We add $\epsilon$ in case there are no labels or predictions in a given behavior class for a video. We also apply a Gaussian blur with standard deviation, $\sigma_b$, and window size, $w_b$, to the groundtruth labels to allow more robustness to small temporal offsets. Our Wasserstein structured loss is then defined as the sum of the cumulative differences over all behaviors:
\begin{equation}
    \begin{array}{ll}
        \mathcal{L}_W = \sum_{b=0}^B \sum_{i=0}^T \left[ \sum_{j=0}^i y_b'^j - \sum_{j=0}^i \hat{y}_b'^j \right]^2
    \end{array}.
\end{equation}
Like the Matching loss, the Wasserstein Loss penalizes extra or missed detections greatly. Because it penalizes differences between the cumulative sums, an extra detection at frame $t$, for example, will introduce a penalty of $T-t$ (Figure~\ref{fig:overview}c). Compared to the Matching Loss, it has the advantage of being fully differentiable, but at the cost of not being able to specify the costs of extra or missed detections precisely. Instead, the cost of these can depend on which frame they occur at within the video. 


\subsection{Per-Frame Loss}
\label{sec:MSELoss}
As for a baseline, we define a per frame loss, $\mathcal{L}_f$, as the mean squared error (MSE) between $Y$ and $\hat{Y}$. We also apply a Gaussian blur to the groundtruth labels $Y$ to allow more robustness to small temporal offsets. 

\begin{equation}
    \begin{array}{ll}
        \mathcal{L}_f = \sum_{b=0}^B \sum_{i=0}^T \left[ y_b^t - \hat{y}_b^t \right]^2
    \end{array}.
\end{equation}

\subsection{Combined Loss}
Similar to~\cite{wei2018sequence}, we found it was helpful to combine the per-frame loss with the structured losses ($\mathcal{L}_H$ and $\mathcal{L}_W$) to improve training.
\begin{equation}
\begin{array}{ll}
\mathcal{L}(Y, \hat{Y}) = \lambda \mathcal{L}_{f}(Y, \hat{Y}) + (1 - \lambda) \mathcal{L}_{s}(Y, \hat{Y})
\label{eq:CombinedLoss}
\end{array}
\end{equation}
where $\mathcal{L}_{f}$ is the per-frame loss, the structured loss $\mathcal{L}_{s}$ is either $\mathcal{L}_W$ or $\mathcal{L}_H$, and $\lambda$ is a hyper parameter between 0 and 1. This is especially useful for the Matching Loss, since there may not be any initial predictions after thresholding the network classifier scores. Additionally, including the per-frame loss helps reduce false negatives for the Matching Loss. We also experimented with decreasing the weight of the per-frame loss during training. 

\section{Visualization}
\begin{figure}[ht]
\centering
\includegraphics[width=0.95\linewidth]{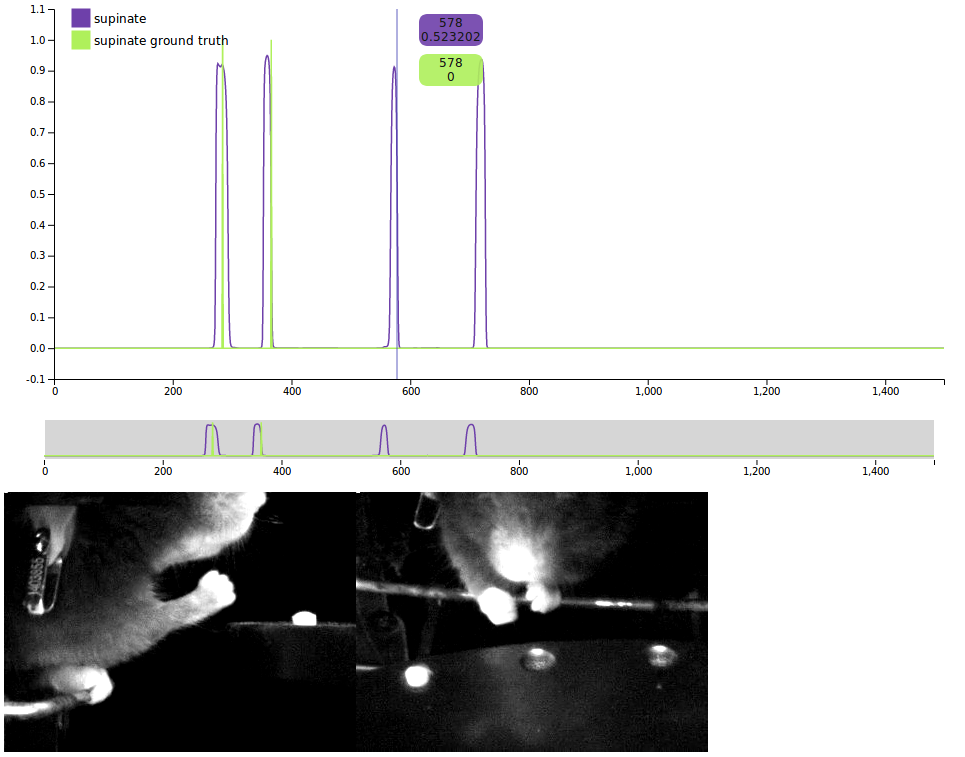}
\caption{An example screen shot of our network-output viewer for videos. Green indicates ground truth and purple the network's predictions. The vertical blue line indicates which video frame is shown below. 
In this example, we investigate the cause of the false positive prediction of ``supinate'' at frame 578. We observe that the mouse has turned its paw, which is part of the supinate behavior, but that it is not near the food pellet.}
\label{fig:viewer}
\end{figure}

We developed visualization code to help understand performance on video data. When debugging various network architectures, we found it difficult to review an individual frame associated with a network prediction and understand what may have caused the network score. In order to help visualize results, we created a web-based viewer (Fig.~\ref{fig:viewer}) that synchronizes the network output score and video frame. The viewer has two main components. A line graph, where the x-axis is video frames and the y-axis is the network output score. The other component is a video viewer, where the frame being shown is the currently selected frame. A frame can be selected by either playing the movie or mousing over the line graph. Being able to slowly, or quickly, mouse over consecutive video frames around curious network outputs helped find software bugs and explore network architectures.

The viewer was created using a JavaScript library called d3.js~\cite{d3js}. Network outputs are stored as a CSV file, where the first column is the frame number and any following columns are any set of scores, such as the ground truth of a label and network predictions for that label. Movie files can be any format supported by HTML5 such as MP4 or WebM. Source code is available at \url{https://github.com/iskwak/DetetctingActionStarts}.

\section{The Mouse Reach Dataset}
\label{sec:dataset}
We introduce the Mouse Reach Dataset, a new video dataset carefully annotated with action bout starts by experts. This dataset was collected by neuroscientists interested in understanding the neural control of behavior, which involves examining recorded neural activity prior to a behavior change~\cite{guo2015cortex}. Unlike most action detection datasets, in which the duration of the bout is labeled, only the action start was relevant. This is because, only action starts, not the action ends, were well-defined and could be labeled consistently (Figure~\ref{fig:overview}). This provides an interesting opportunity for computer vision research to develop tools to automatically detect action starts for neuroscience research. The dataset is available at \url{http://research.janelia.org/bransonlab/MouseReachData/}.

The dataset contains 1165 recordings of four mice attempting to grab and eat a food pellet. Only the mice's limbs are free to move. They were recorded many times a day for several days from two fixed, time-synchronized cameras. The  videos were recorded at 500 frames per second in near infrared. 

The biologists labeled the start of six different actions. ``Lift'' occurs when the mouse begins to lift its paw from the perch. ``Hand-open'' occurs when the paw begins to open before grabbing the pellet. ``Grab'' occurs when the paw begins to close around the pellet. ``Supinate'' occurs when the paw begins to turn toward the mouse's mouth. ``At-mouth'' occurs when the pellet is first placed in the mouse's mouth. ``Chew'' occurs when the mouse begins to eat the pellet. Note that Grab and Supinate are also annotated when the mouse fails to grab the pellet. The most common behavior is the ``Hand-open'' behavior with 2227 labels -- on average 1.91 labeled instances per video. The least common behavior is ``Chew'', with 664 labeled instances (see supplementary materials for details).

In these videos, the temporal sequence of actions is highly structured. For example, the mouse cannot chew a food pellet without grabbing it first. Thus, long-range temporal context is important. The most common sequence of labels is ``Lift'', ``Hand-open'', ``Grab'', ``Supinate'', ``At-mouth'', and ``Chew''. However, the temporal sequence is also flexible, as when the mouse misses a pellet it will backtrack and repeat actions. For example, multiple instances of grab can occur in a row. This also produces an imbalance in action categories.

We believe this dataset will provide computer vision researchers an opportunity to work with high quality labels of action key frames. As mentioned previously, bout boundary detection has gained interest in the vision research community, and this provides a dataset for comparing and spurring algorithm development while providing useful tools for neuroscientists.
\section{Experiments}
\label{sec:experiments}
The following sections describes our experimental setups on the Mouse Reach and THUMOS'14 datasets. Source code is available at \url{https://github.com/iskwak/DetetctingActionStarts}.

\begin{table*}
\centering
\begin{small}
\begin{tabular}{|l|l|l|l|l|l|l|}
\hline
Algorithm                   & Lift          & Hand-open     & Grab          & Supinate      & At-mouth & Chew \\ \hline
MSE+HOGHOF                  & 0.79          & 0.78          & \textbf{0.84} & 0.65          & 0.44          & 0.45 \\ \hline
Matching+HOGHOF             & 0.88          & 0.77          & \textbf{0.84} & \textbf{0.75} & 0.43          & 0.45 \\ \hline
Wasserstein+HOGHOF          & \textbf{0.91} & 0.78          & 0.83          & 0.73          & 0.49          & \textbf{0.46} \\ \hline
MSE+Canned I3D              & 0.81          & 0.77          & 0.83          & 0.68          & 0.46          & 0.41 \\ \hline
Matching+Canned I3D         & 0.83          & 0.74          & 0.80          & 0.70          & 0.44          & 0.33 \\ \hline
Wasserstein+Canned I3D      & 0.83          & 0.74          & 0.80          & 0.71          & 0.44          & 0.33 \\ \hline
MSE+Finetuned I3D           & 0.61          & 0.51          & 0.53          & 0.45          & 0.28          & 0.29 \\ \hline
Matching+Finetuned I3D      & 0.90          & 0.80          & \textbf{0.84} & 0.74          & \textbf{0.51} & 0.35 \\ \hline
Wasserstein+Finetuned I3D   & 0.88          & \textbf{0.81} & \textbf{0.84} & 0.72          & 0.48          & 0.26 \\ \hline
Finetuned I3D+Feedforward   & 0.38          & 0.25          & 0.37          & 0.30          & 0.17          & 0.15 \\ \hline
ODAS                        & 0.35          & 0.45          & 0.59          & 0.40          & 0.27          & 0.13 \\ \hline
\end{tabular}
\end{small}
\caption{F1 scores for each loss, feature type, and behavior. Matching and Wasserstein losses outperform the per-frame MSE. }
\label{table:behavresults}
\end{table*}

\begin{table}
\centering
\begin{tabular}{|l|l|l|l|l|l|l|}
\hline
 Algorithm                  & F1 Score      & Precision     & Recall\\ \hline
 MSE+HOGHOF                 & 0.69          & 0.62          & 0.79 \\ \hline
 Matching+HOGHOF            & 0.73          & 0.72          & 0.74  \\ \hline
 Wasserstein+HOGHOF         & \textbf{0.75} & 0.78          & 0.71  \\ \hline
 MSE+Canned I3D             & 0.70          & 0.64          & 0.76 \\ \hline
 Matching+Canned I3D        & 0.69          & 0.72          & 0.66 \\ \hline
 Wasserstein+Canned I3D     & 0.73          & 0.77          & 0.70  \\ \hline
 MSE+Finetune I3D           & 0.48          & 0.37          & 0.69 \\ \hline
 Matching+Finetune I3D      & \textbf{0.75} & 0.80          & 0.70 \\ \hline
 Wasserstein+Finetune I3D   & \textbf{0.75} & \textbf{0.85} & 0.66 \\ \hline
 I3D+Feedforward            & 0.27          & 0.16          & 0.88 \\ \hline
 ODAS                       & 0.22          & 0.12          & \textbf{0.94} \\ \hline
\end{tabular}
\caption{F1 score, precision, and recall for each loss and feature type, combining all behaviors. The Matching and Wasserstein Losses have better F1 score and precision than MSE, implying fewer false positives.}
\label{table:fullresults}
\end{table}

\subsection{Mouse Experiments}
We test our loss functions on the Mouse Reach Dataset. The goal of this task is to detect the start of a behavior within $\tau$ frames. For these experiments we set $\tau=10$ frames (0.02 seconds), based on a study of annotation consistency on a subset of the videos. For each of the four mice, we train with all other mice's videos and the first half of that mouse's videos, then test on the second half of that mouse's videos. We included training videos from the test mouse because, without this, for all algorithms tested, generalization across mice was poor. Test sets consisted of 125, 55, 274, and 192 videos for the four mice.


\subsection{THUMOS'14 Experiments}
The most similar work to ours is ODAS~\cite{shou2018online}, which detects action starts in an online setting. Although our work is not designed to be used in an online setting, we can use their evaluation protocol to test our algorithm on a known action detection dataset, THUMOS'14~\cite{THUMOS14}. \cite{shou2018online} defines point-level average precision, p-AP, as a way to evaluate the performance of start frame detection. For each class, the predicted action starts are ordered by their confidences. Each prediction is counted as a correct action start if it matches the ground truth action start class, the temporal distance between the prediction and true start is less than a threshold, and no other prediction has been matched to this start. With the matches, we can compute the point-level average precision, p-AP, and the average over classes to compute the p-mAP.

Following \cite{shou2018online}, we use two metrics to evaluate our algorithm on THUMOS'14. The first is p-mAP, computed over different temporal distance thresholds. The temporal offsets are set at every second from 1 to 10 seconds. These offsets can provide insight on how precise an action start can be detected, which is useful for tuning algorithms to different applications. The second is \textit{AP depth at recall X\%} which is the average of precision points on the P-R curve between 0\% and X\% recall. The \textit{AP depth at recall X\%} can be useful to evaluate the precision of top predictions at low recall.

\begin{figure}[htb]
\begin{center}
\begin{tabular}{ccc}
 -10 frames & Annotated start & +10 frames\\ 
\includegraphics[width=0.3\linewidth]{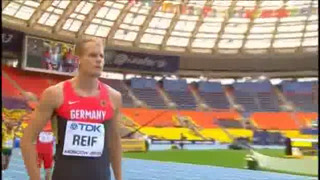} &
\includegraphics[width=0.3\linewidth]{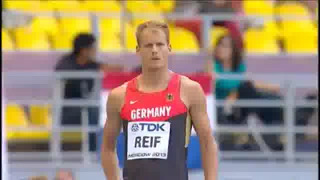} &
\includegraphics[width=0.3\linewidth]{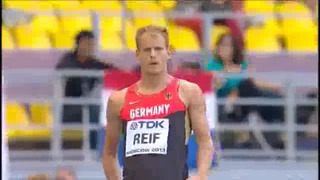} \\ 
\includegraphics[width=0.3\linewidth]{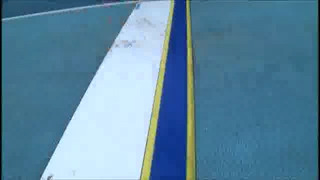} &
\includegraphics[width=0.3\linewidth]{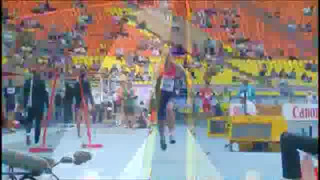} &
\includegraphics[width=0.3\linewidth]{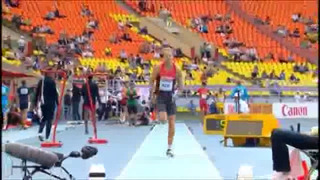} \\ 
\end{tabular}
\end{center}
\caption{Examples of labeled from THUMOS'14 showing inconsistency in start  annotations. Each row shows the annotated starts (middle), 10 frames before (left), and 10 frames after (right) for different long jump examples. The annotated starts do not correspond to the same behavior. In the top row, the annotated start is when the athlete leans back before running. In the bottom row, the example is a clipped replay, and the annotated start is after the athlete has already started his run. While the per-frame annotations are consistent, because the video only shows a portion of the long jump, the derived starts are inconsistent. In contrast, in our Mouse Reach Dataset, the start frames are explicitly annotated to correspond to the same behavior.}
\label{fig:thumos_difficulties}
\vspace{-0.4cm}
\end{figure}

Although THUMOS'14 has been used as a baseline for online action start detection, we feel that our dataset is better suited for action start detection. In Figure~\ref{fig:thumos_difficulties}, we show example frames of action starts of two behaviors. In each case the start of the behavior is not clearly seen. Because THUMOS'14 is labeled for fine-grained action detection, the start of a bout may not represent the start of the action. Some labeled sequences are of highlights of the action and only show a subset of the action. For example, a long jump action clip may only show the portion of the action where the athlete jumps into the sand pit. However we believe the start of the action should be when the athlete begins their run towards the sand pit.

\setlength{\tabcolsep}{3pt} 
\setlength{\tabcolsep}{6pt} 

\subsection{Experimental Details}
For all experiments, we used pre-computed or fine-tuned features as inputs to an RNN. Our base model is a two layer bi-directional LSTM with 256 hidden units. The inputs to the LSTM pass through a fully connected layer, ReLU, and Batch Normalization. The outputs are transformed by a fully connected layer with a sigmoid activation layer. A figure of our model is available in the supplementary material. We used ADAM for optimization. The learning rate was tuned for each loss function. The network was trained for 400 epochs with a batch size of 10.


For the Mouse Reach dataset experiments we used two types of input feature: HoG+HOF~\cite{dalal2005histograms} and I3D~\cite{carreira2017quo}. The HoG+HOF are hand-designed features that capture image gradients and motion gradients. We computed these features on overlapping windows on each view point, resulting in an 8000 dimensional feature vector. I3D is a state-of-the-art action recognition network that uses sequences of RGB and optical flow frames as input. We used the output of the last average pooling layer before the $1\times1\times1$ convolutional classification layer as the I3D feature representation. For each frame in the video sequence, I3D was applied to a 64 frame window, centered around the input frame. We refer to the features from a model\footnote{https://github.com/deepmind/kinetics-i3d} pre-trained on the Kinetics dataset~\cite{kay2017kinetics} as Canned I3D. We also fine-tune the I3D network on our dataset by training the feedforward, per-frame I3D network on the Mouse Reach Dataset. This feature set is referred to as Finetuned I3D. HoG+HOF features are provided with the dataset.

We trained RNNs with each of the three feature types (HOGHOF, Canned I3D, and Finetuned I3D) and each of three losses: Matching (Section~\ref{sec:MatchingLoss}), Wasserstein (Section~\ref{sec:WassersteinLoss}), and MSE (Mean-squared error, Section~\ref{sec:MSELoss}).

When using the HOG+HOF and Canned I3D features as inputs with the Matching $\mathcal{L}_H$ in the combined loss (\ref{eq:CombinedLoss}) the weight of the per-frame loss $\lambda$ was reduced from an initial value $0.99$ to $0.5$ with an exponential step size of $0.9$ every five epochs until $\mathcal{L}_f$ and $\mathcal{L}_s$ were weighted equally. For the Finetuned I3D features, $\lambda$ was reduced to $0.25$. We set the $C_{tp} = 4$, $C_{fp} = 1$, $C_{fn} = 2$. For the Wasserstein Loss, $\lambda = 0.5$. In order to help all our losses deal with the scarcity of positive samples, we blur the ground truth label sequence with a Gaussian kernel with window size 19 frames and standard deviation of 2 frames. Like our choice of $\tau$, the window size was chosen based off of our annotation consistency study.

For the THUMOS'14 experiments we created features from an I3D network pre-trained on the Kinetics Dataset~\cite{carreira2017quo}. Like the Mouse Reach experiments, we used outputs before the classification layer as a feature representation. We only used the pre-trained RGB 3D convolutional branch of the I3D network. We used the same hyper parameters for our losses as the Mouse Reach experiments, except the maximum matching offset was $\tau=30$ frames, 1 second. In the mouse videos, the researchers were interested in start detections within milliseconds of the behavior starting, however for THUMOS'14, we are evaluating results at the level of seconds. For the same reasons we blur the ground truth sequence with a kernel window size of 59 frames and standard deviation of 8 frames. In addition to our bi-directional LSTM, we train a forward only LSTM. A forward only LSTM could be used in an online fashion, which should be more comparable to the results from \cite{shou2018online}.

\begin{figure}[tb]
\centering
\begin{tabular}{lllllll}
\includegraphics[width=0.90\linewidth]{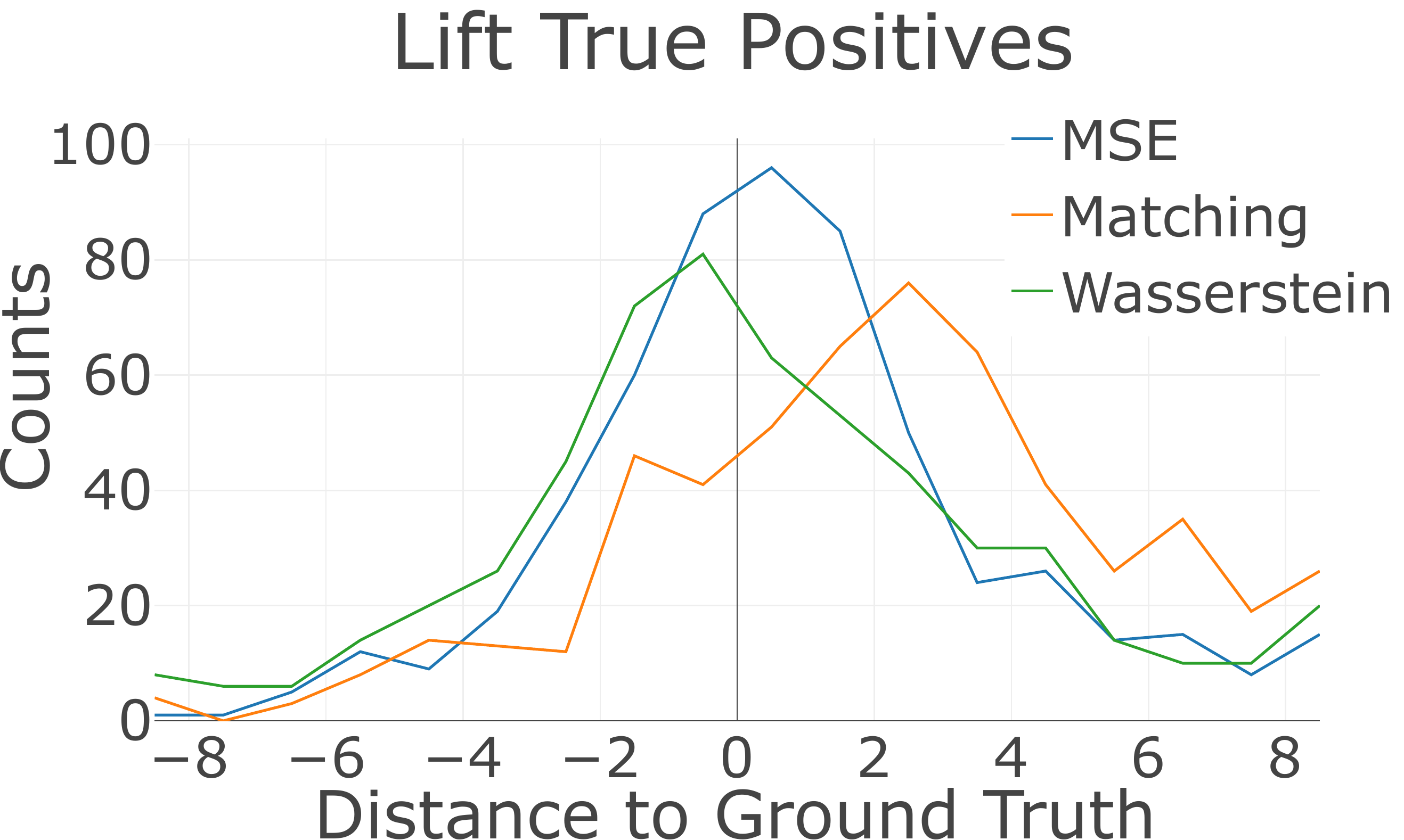} \\ \includegraphics[width=0.90\linewidth]{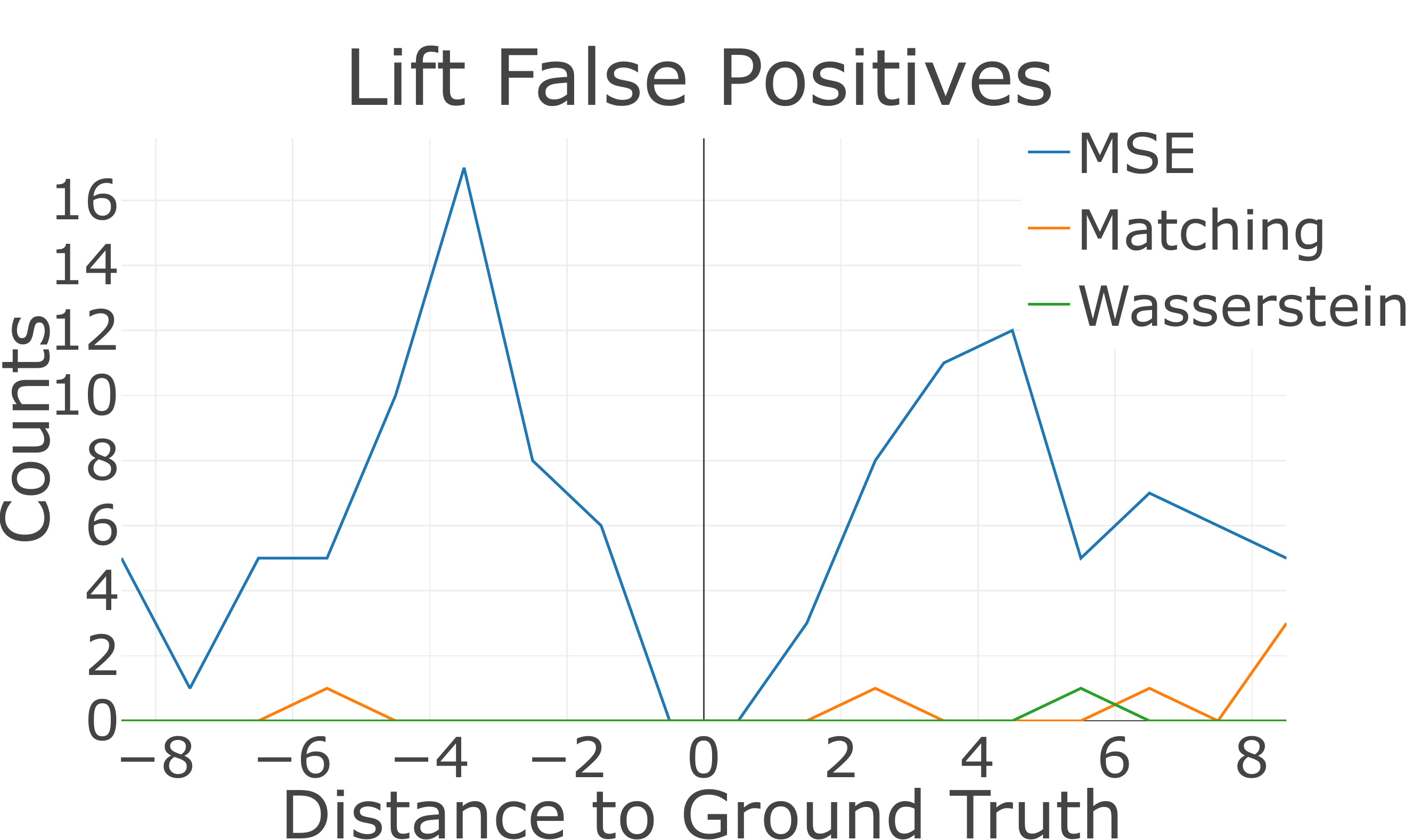} \\
\end{tabular}
\caption{Histogram of offsets of lift detections from closest groundtruth starts for true positives (top) and false positives (bottom). Many false positives are more than 10 frames from any true positive, and are not counted here. For true positives, the majority of detections have a very small offset. Histograms for other behaviors are in the supplementary materials.}
\label{fig:fpstps}
\end{figure}

\subsection{Mouse Reach Results}
We consider a network trained with MSE loss as a baseline and compare it to a networks trained with our two structured loss functions. RNNs trained with the MSE loss are equivalent to the most standard framework for action detection, if the bout lengths are just one frame. Table~\ref{table:fullresults} shows the precision, recall and F1 score for each of the proposed loss functions and feature representations. We also show results of the fine-tuned feed-forward I3D network. The structured losses have a better F1 score because they have higher precision, implying an improved false positive rate. This is one of the goals of our structured losses, to penalize spurious start detections. The Matching loss explicitly penalizes false positive predictions and the Wasserstein Loss attempts to match the number of predicted behavior starts with the ground truth. Overall, the Wasserstein Loss performs best regardless of the input features. In Table~\ref{table:behavresults}, we can see the breakdown performance with respect to each behavior. The action categories that benefited most from the structured losses were lift and supinate. Perhaps surprisingly, we do not see a big difference in performance between HOG/HOF features and the fine-tuned I3D features learned for this task.

\begin{table*}
\centering
\begin{small}
\begin{tabular}{|l|l|l|l|l|l|l|l|l|l|l|}
\hline
Offsets  & 1 & 2 & 3 & 4 & 5 & 6 & 7 & 8 & 9 & 10 \\ \hline
SceneDetect & 0.01 & 0.02 & 0.02 & 0.03 & 0.03 & 0.04 & 0.04 & 0.05 & 0.05 & 0.05 \\ \hline
ShotDetect & 0.01 & 0.02 & 0.02 & 0.03 & 0.03 & 0.03 & 0.04 & 0.04 & 0.04 & 0.05 \\ \hline
ODAS \cite{shou2018online} & 0.03 & 0.04 & 0.04 & 0.05 & 0.05 & 0.06 & 0.06 & 0.07 & 0.07 & 0.08 \\ \hline
MSE-FWD & 0.06 & 0.10 & 0.11 & 0.12 & 0.13 & 0.14 & 0.15 & 0.15 & 0.15 & 0.15 \\ \hline
Matching-FWD & 0.06 & 0.10 & 0.11 & 0.12 & 0.13 & 0.14 & 0.15 & 0.15 & 0.15 & 0.15 \\ \hline
Wasserstein-FWD & 0.05 & 0.09 & 0.10 & 0.11 & 0.12 & 0.12 & 0.13 & 0.13 & 0.13 & 0.13 \\ \hline
MSE+BIDIR & 0.08 & \textbf{0.14} & 0.15 & 0.16 & 0.16 & 0.17 & 0.18 & 0.18 & 0.18 & 0.18 \\ \hline
Matching+BIDIR & \textbf{0.09} & \textbf{0.14} & \textbf{0.16} & \textbf{0.17} & \textbf{0.18} & \textbf{0.19} & \textbf{0.19} & \textbf{0.20} & \textbf{0.20} & \textbf{0.20} \\ \hline
Wasserstein+BIDIR & 0.05 & 0.09 & 0.11 & 0.12 & 0.12 & 0.13 & 0.13 & 0.14 & 0.14 & 0.14 \\ \hline
\end{tabular}
\end{small}
\caption{p-mAP at depth \textit{Rec}=1 shows the performance of our proposed loss functions on THUMOS'14 at different offset thresholds. The *+FWD networks were trained as forward only LSTM's, whereas the *+BIDIR networks were bi-directional LSTM's.}
\label{table:pmapthumos}
\end{table*}

\begin{figure}[htb]
\centering
\begin{small}
\begin{tabular}{lllllll}
\includegraphics[width=0.95\linewidth]{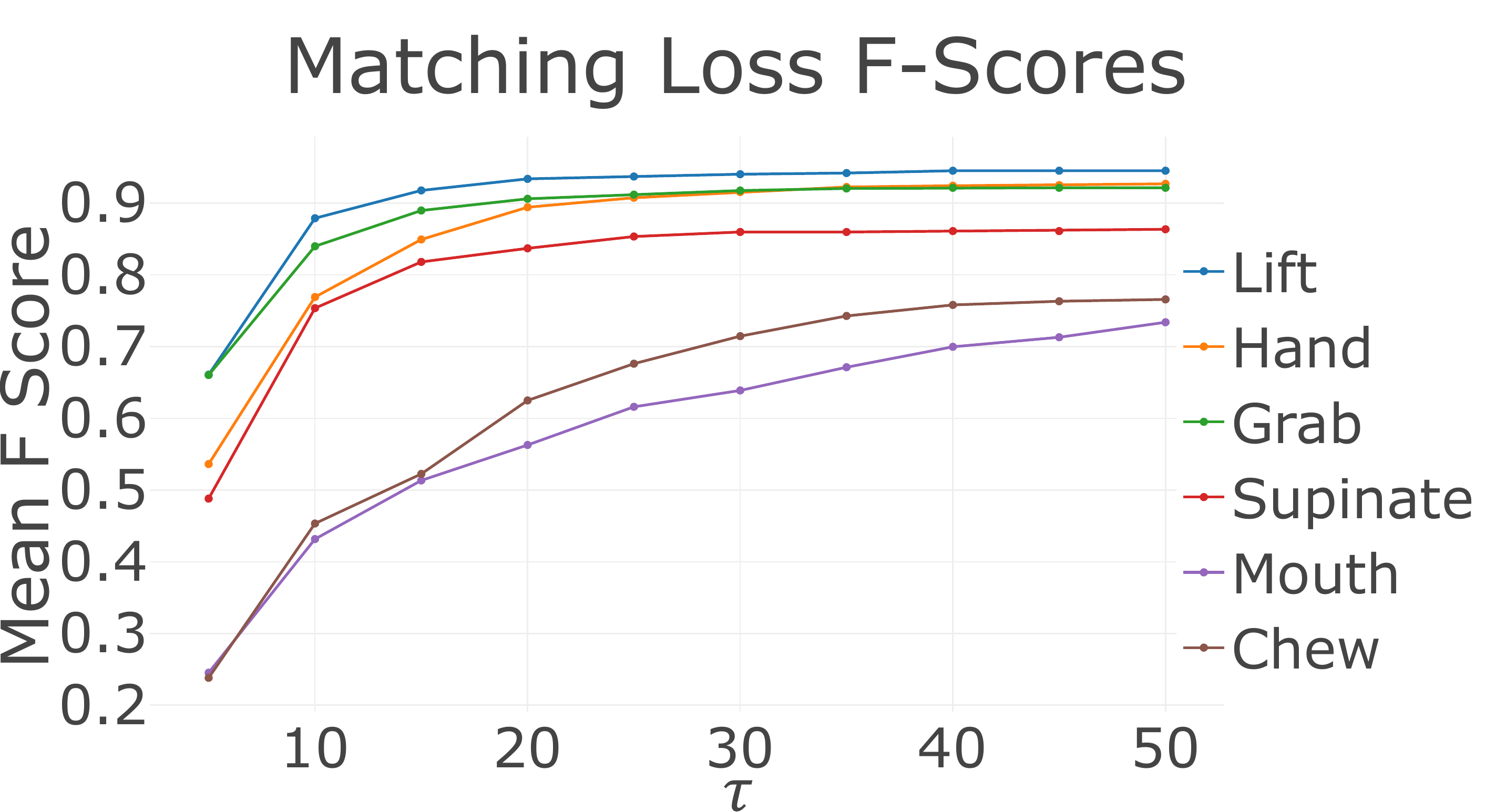} \\ \includegraphics[width=0.95\linewidth]{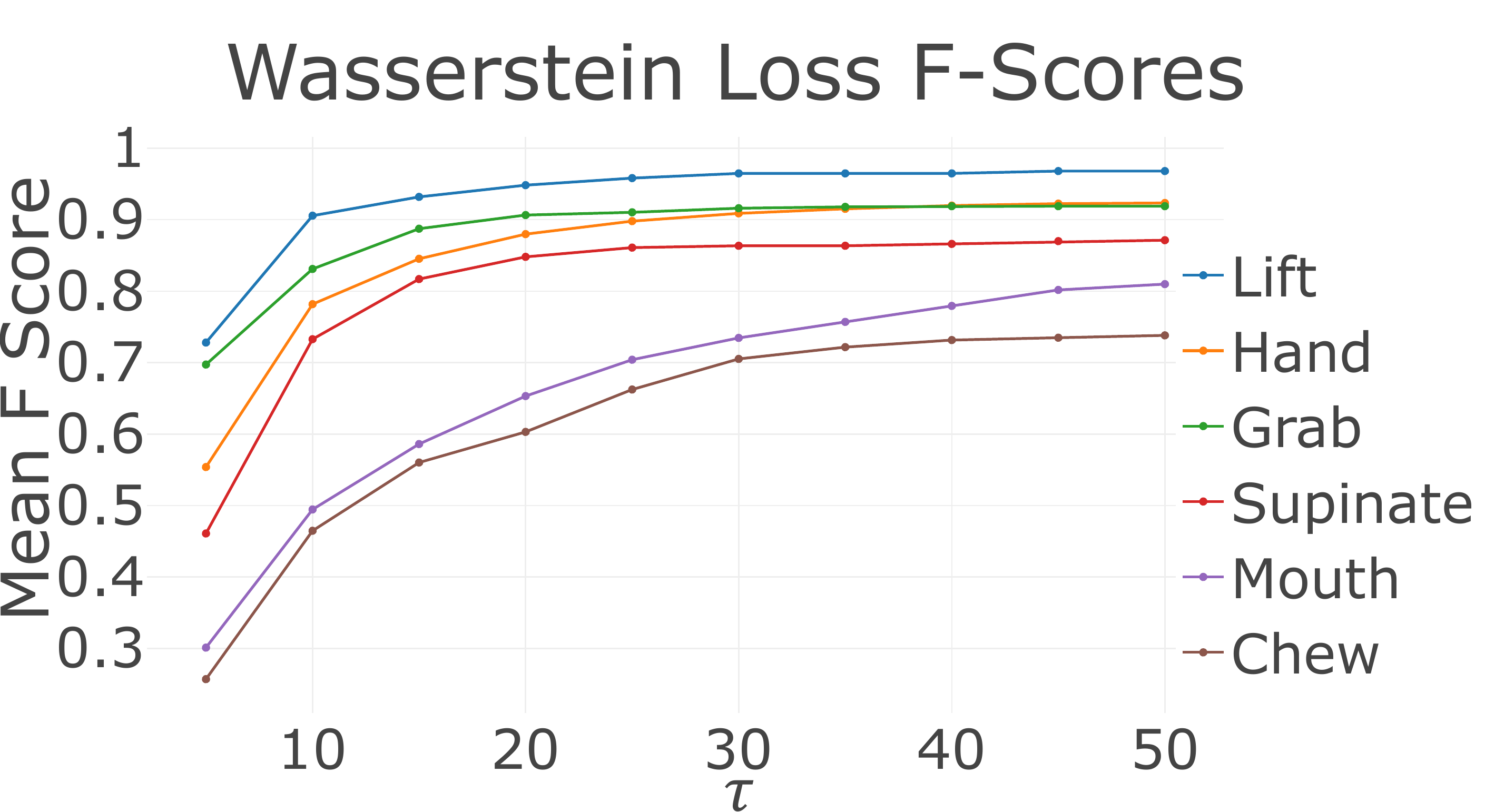}
\end{tabular}
\end{small}
\caption{F1 score as we vary $\tau$, the threshold on frame-distance for a correct match, for Matching Loss (top) and Wasserstein Loss (bottom). Note We do not re-train any networks, only reanalyze the network predictions. For the first 4 behaviors, F1 score plateaus after $\tau = 20$. the network detects the start within 20 frames from the true start. At-mouth and Chew benefit from a much larger $\tau$.}
\label{fig:varyfscore}
\end{figure}

\begin{table*}
\centering
\begin{small}
\begin{tabular}{|l|l|l|l|l|l|l|l|l|l|l|}
\hline
Depth@X  & 0.1 & 0.2 & 0.3 & 0.4 & 0.5 & 0.6 & 0.7 & 0.8 & 0.9 & 1.0 \\ \hline
SceneDetect & 0.30 & 0.18 & 0.12 & 0.09 & 0.07 & 0.06 & 0.05 & 0.04 & 0.4 & 0.03 \\ \hline
ShotDetect & 0.26 & 0.15 & 0.11 & 0.08 & 0.07 & 0.06 & 0.05 & 0.04 & 0.04 & 0.03 \\ \hline
ODAS \cite{shou2018online} & 0.42 & 0.27 & 0.19 & 0.14 & 0.11 & 0.10 & 0.08 & 0.07 & 0.06 & 0.05 \\ \hline
MSE+FWD & 0.55 & 0.48 & 0.46 & 0.45 & 0.45 & 0.45 & 0.44 & 0.44 & 0.44 & 0.44 \\ \hline
Matching+FWD & \textbf{0.64} & \textbf{0.57} & \textbf{0.55} & \textbf{0.54} & \textbf{0.54} & \textbf{0.53} & \textbf{0.53} & \textbf{0.53} & \textbf{0.53} & \textbf{0.53} \\ \hline
Wasserstein+FWD & 0.58 & 0.55 & 0.53 & 0.53 & 0.53 & \textbf{0.53} & \textbf{0.53} & \textbf{0.53} & \textbf{0.53} & \textbf{0.53} \\ \hline
MSE+BIDIR & 0.46 & 0.36 & 0.30 & 0.26 & 0.24 & 0.24 & 0.23 & 0.23 & 0.23 & 0.23 \\ \hline
Matching+BIDIR & 0.59 & 0.52 & 0.48 & 0.45 & 0.44 & 0.43 & 0.42 & 0.42 & 0.42 & 0.42 \\ \hline
Wasserstein+BIDIR & 0.42 & 0.35 & 0.33 & 0.31 & 0.30 & 0.29 & 0.29 & 0.29 & 0.29 & 0.29 \\ \hline
\end{tabular}
\end{small}
\caption{Average p-mAP at different depths on the THUMOS'14 dataset.}
\label{table:avgpmapthumos}
\end{table*}

The MSE+Finetuned I3D performs far worse than we expected. We believe that improved sampling of negatives, e.g. hard negative mining, could correct this. We used labeled starts as positives, frames more than 10 frames from starts as negatives, and do not include other frames within 10 frames of action starts. We suspect predictions from this network near starts are also positive, resulting in many false positives.

Figure~\ref{fig:fpstps} shows the distribution of predictions for the lift (other behaviors shown in the supplementary materials). Within the $\tau=10$ frames of the true start, we see that our structured losses predict far fewer false positives without missing true positives. Thus, our losses are more likely to produce a single prediction for an action start than MSE. Figure~\ref{fig:varyfscore} shows how predictions far from the action start eventually are matched. 

At the time of this writing, source code and the trained model for the ODAS~\cite{shou2018online} algorithm was unavailable, so we reimplemented their algorithm as best we could. To make comparisons as fair as possible, we used I3D as the backbone network. The network was trained with adaptive sampling and temporal consistency, but without a GAN to generate hard negatives. Our implementation of ODAS without a GAN model was slightly better than the I3D+Feedforward model. However our recurrent models all outperform our implementation of ODAS.

\subsection{THUMOS'14 Results}
Table~\ref{table:pmapthumos} shows the p-mAP performance of our algorithms with different offset thresholds compared to ODAS and baselines provided in~\cite{shou2018online}. The Matching+BIDIR algorithm equals or outperforms all other algorithms for each threshold offset. Interestingly, the Wasserstein+BIDIR loss performs worse then both MSE+BIDIR and Matching+BIDIR on THUMOS'14. Across each cost function, using bidirectional networks provides about a 0.02 improvement at most threshold offsets. For future work we will investigate forward only for the Mouse Reach dataset.

Our baseline MSE+FWD algorithm performs better than ODAS. It's possible that our label re-weighting is causing our baseline forward algorithms to perform better than ODAS. As mentioned previously label balance was improved through sampling positive samples more often. Applying our label re-weighting may help in this example as well.

The average p-mAP \textit{at depth X\%} results are shown in Table~\ref{table:avgpmapthumos}. All of our algorithms perform better than ODAS at each depth. Interestingly, the FWD only algorithms perform worse at each offset threshold, but perform better on the average p-mAP at each depth.

\section{Conclusion}
In this work we show that it is possible to predict the frame where behaviors start with high accuracy. Due to the nature of our task, we focused on developing structured loss functions that would reduce the number of false positive predictions. For the Mouse Reach Dataset we found the Wasserstein loss to be easier to optimize and performed better. However for the THUMOS'14 dataset the Matching loss performed the best. In the future we plan on testing the importance of $\tau$ and smoothing the ground truth labels on the performance of our loss functions. By modifying $\tau$ or the size of the Gaussian kernel on the ground truth labels, we can adjust the importance of predicting the action start precisely.

{\small
\bibliographystyle{ieee}
\bibliography{egbib}
}

\end{document}


\title{Supplementary Materials}


\maketitle
\ifwacvfinal\thispagestyle{empty}\fi

\begin{table}
\centering
\begin{tabular}{|c|c|c|}
\hline
 Behavior & Total & Average Per Video \\ \hline
 Lift        & 1175  & 1.01              \\ \hline
 Hand-open        & 2227  & 1.91              \\ \hline
 Grab        & 2096  & 1.79              \\ \hline
 Supinate    & 1392  & 1.19              \\ \hline
 At-mouth       & 921   & 0.79              \\ \hline
 Chew        & 664   & 0.57              \\ \hline
 Background  & 830939 & 71081            \\ \hline
\end{tabular}
\caption{Number of labelled frames with the Mouse Reach Dataset.}\label{table:frames}
\end{table}

\begin{table}
\centering
\begin{tabular}{|c|c|}
\hline
 Mouse    & Total Videos \\ \hline
 M134     & 217          \\ \hline
 M147     & 97           \\ \hline
 M173     & 492          \\ \hline
 M174     & 359          \\ \hline
\end{tabular}
\caption{The Mouse Reach Dataset contains a total of 1169 videos of mice performing the reaching task.}\label{table:videos}
\end{table}

\begin{figure*}[ht]
\centering
\begin{tabular}{lllllll}
\includegraphics[width=0.45\linewidth]{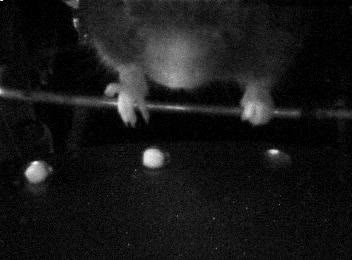} & \includegraphics[width=0.45\linewidth]{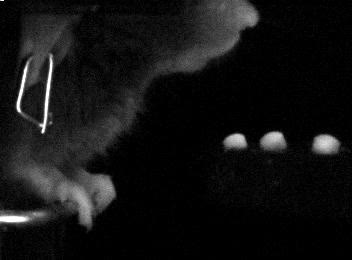} \\
\includegraphics[width=0.45\linewidth]{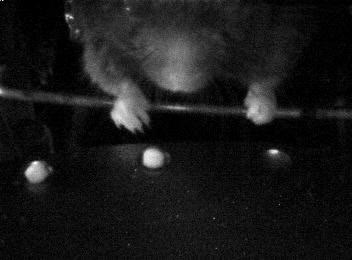} & \includegraphics[width=0.45\linewidth]{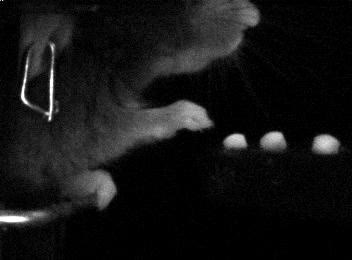} \\
\includegraphics[width=0.45\linewidth]{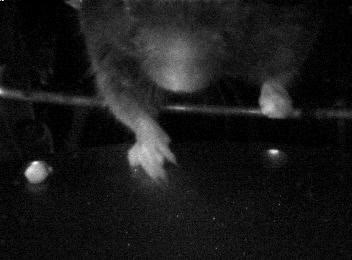} & \includegraphics[width=0.45\linewidth]{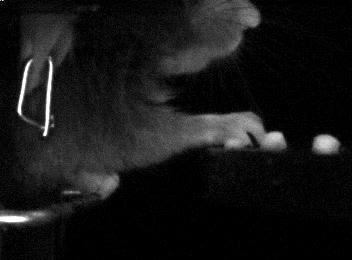} \\
\end{tabular}
\caption{Example frames of behaviors the Mouse Reach Dataset. The first row is the Lift behavior. Here the mouse paw is beginning to move off of the perch. The next row is the Hand-open behavior. Here is the mouse beginning to open his paw to grab a pellet. The third row is the Grab behavior. The mouse beginning to close his paw around a food pellet.}
\label{fig:lifthandgrab}
\end{figure*}

\begin{figure*}[ht]
\centering
\begin{tabular}{lllllll}
\includegraphics[width=0.45\linewidth]{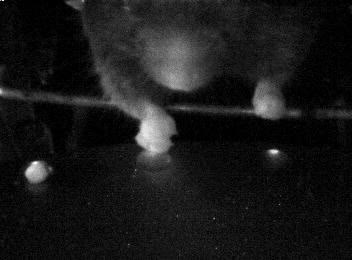} & \includegraphics[width=0.45\linewidth]{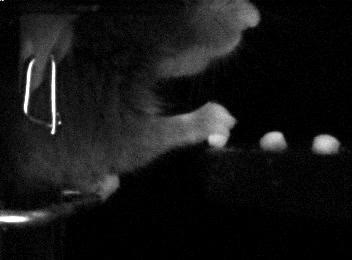} \\
\includegraphics[width=0.45\linewidth]{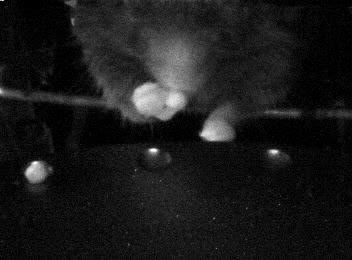} & \includegraphics[width=0.45\linewidth]{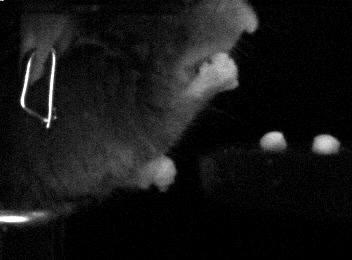} \\
\includegraphics[width=0.45\linewidth]{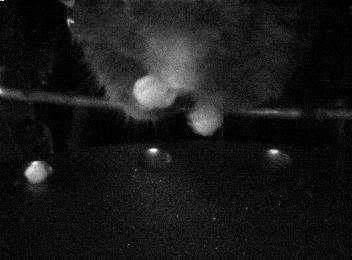} & \includegraphics[width=0.45\linewidth]{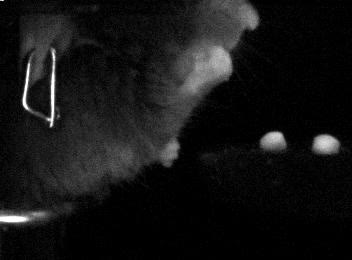}\\
\end{tabular}
\caption{The Supinate behavior is shown in the first row. The mouse is beginning to turn its paw towards its mouth. The second row shows the At-mouth behavior. The mouth behavior occurs when the food pellet is starting to be placed into the mouth. The last row shows the Chew behavior, where the food pellet in the mouth and the mouse is starting to eat the pellet.}
\label{fig:supmouthchew}
\end{figure*}

Table~\ref{table:frames} and Table~\ref{table:videos} provide more details on our dataset. The behaviors: Hand, Grab and Supinate, occur more often because the mouse will fail to grab the food pellet and try to grab food pellet again. The number of chew frames are low because the mouse will also fail to eat the food pellet. Figure~\ref{fig:lifthandgrab} and Figure~\ref{fig:supmouthchew} show sample frames of each of the behaviors.

\begin{figure*}[ht]
\centering
\begin{tabular}{lllllll}
\includegraphics[width=0.9\linewidth]{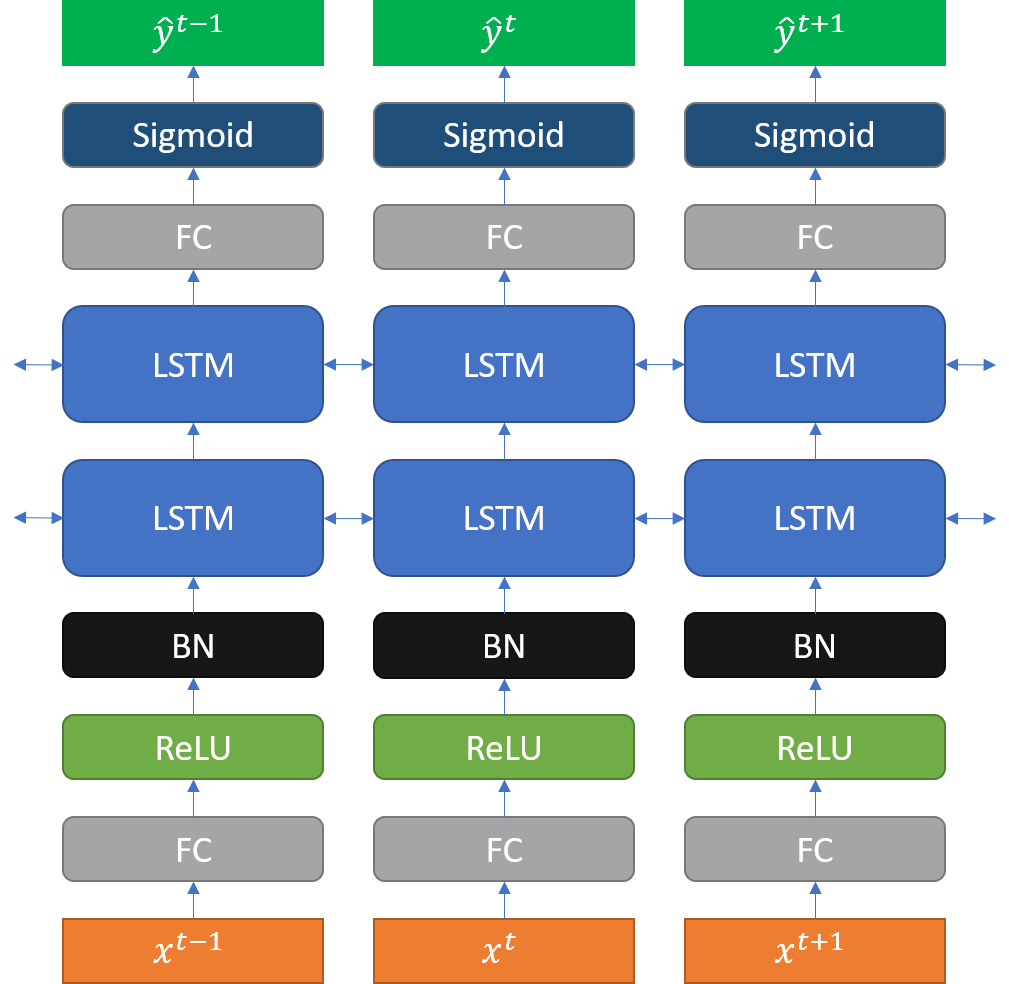} 
\end{tabular}
\caption{Our complete model consists of a fully connected layer, ReLU, Batch Normalization, two Bi-directional LSTM layers, a fully connected layer then a sigmoid activation layer. The LSTMs each have 256 hidden units.}
\label{fig:networkmodel}
\end{figure*}

Fig.~\ref{fig:networkmodel} shows a diagram of our base model. The base model is a two layer bi-directional LSTM with 256 hidden units. The inputs to the LSTM pass through a fully connected layer, ReLU, and Batch Normalization. The outputs are transformed by a fully connected layer with a sigmoid activation layer.

\begin{figure*}[ht]
\centering
\begin{tabular}{lllllll}
\includegraphics[width=0.45\linewidth]{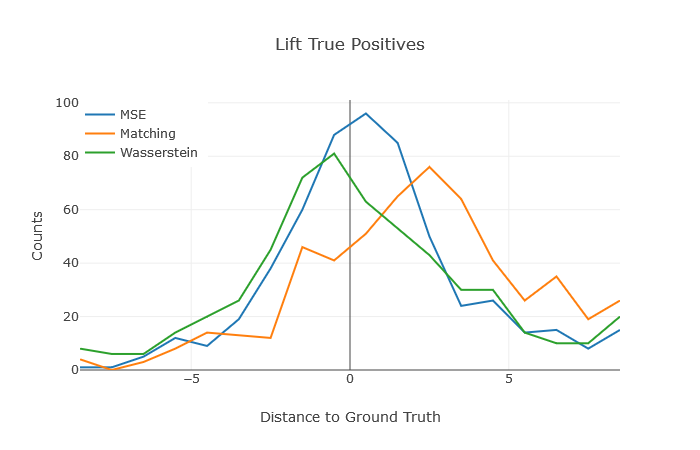} & \includegraphics[width=0.45\linewidth]{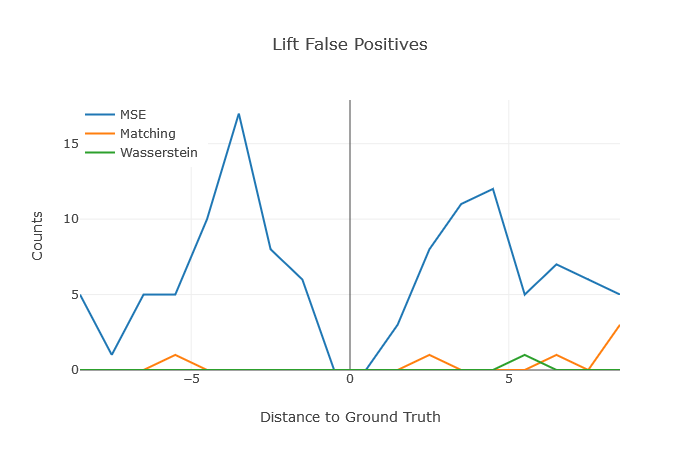} \\
\includegraphics[width=0.45\linewidth]{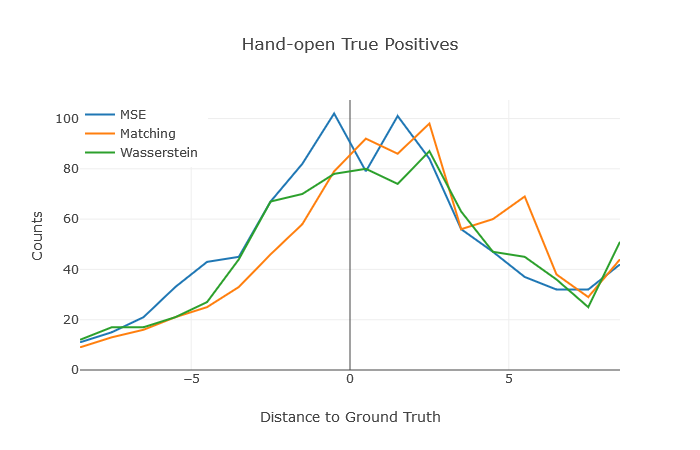} & \includegraphics[width=0.45\linewidth]{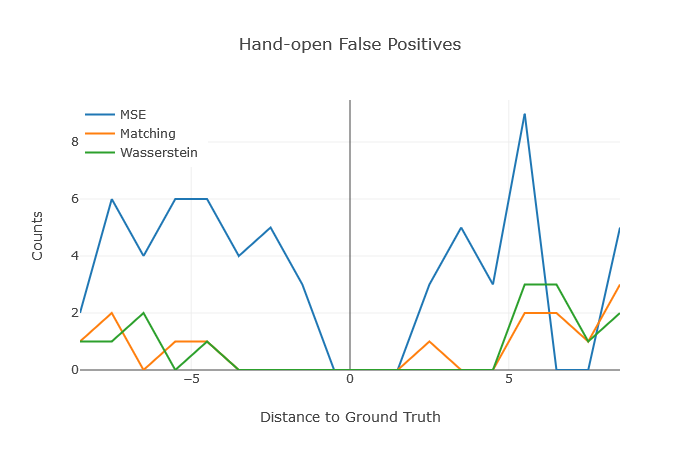} \\
\includegraphics[width=0.45\linewidth]{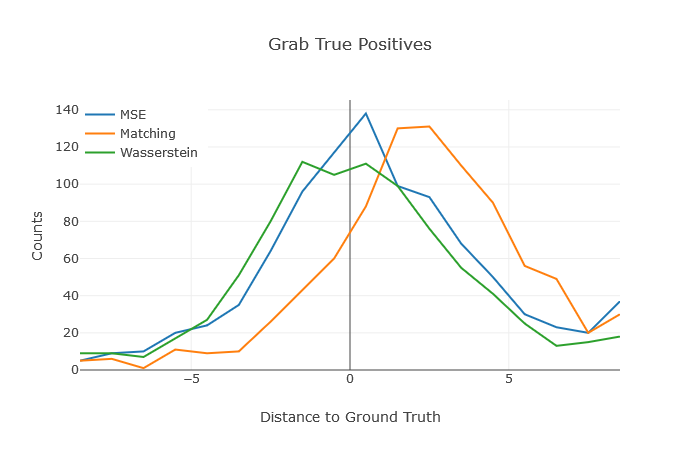} & \includegraphics[width=0.45\linewidth]{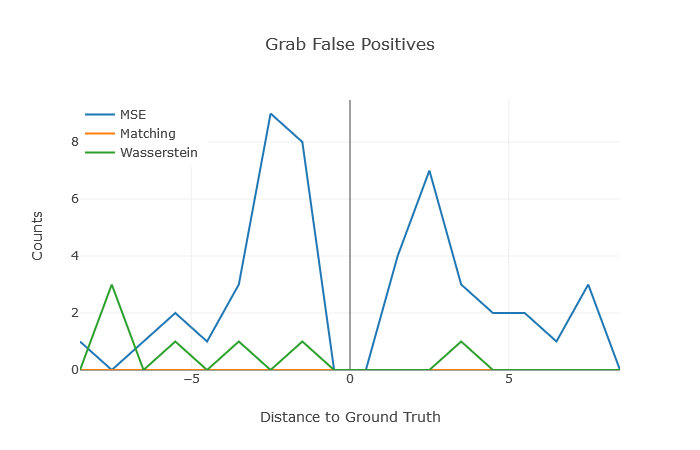} \\
\end{tabular}
\caption{Left column shows the distribution of true positives and the right side the false positives. For these behaviors the network is able to localize the start frame accurately.}
\label{fig:fpstps1a}
\end{figure*}

\begin{figure*}[ht]
\centering
\begin{tabular}{lllllll}
\includegraphics[width=0.45\linewidth]{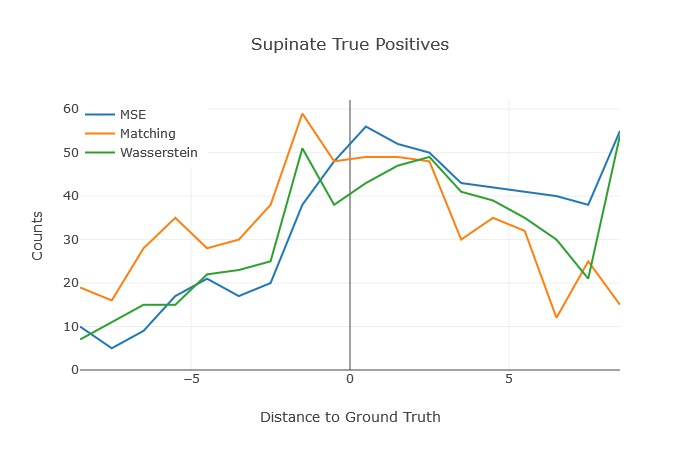} & \includegraphics[width=0.45\linewidth]{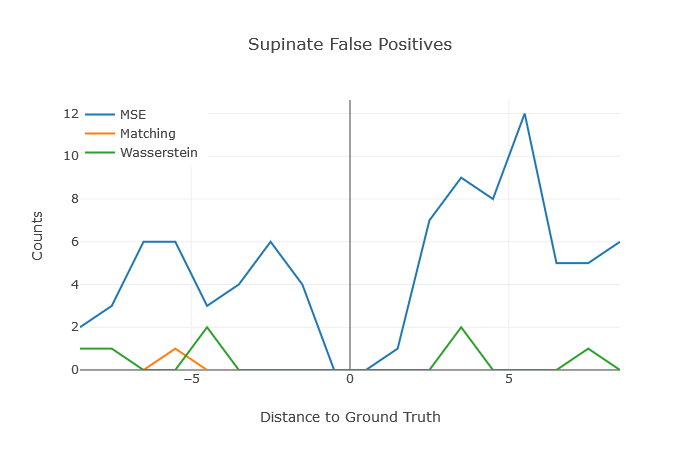} \\
\includegraphics[width=0.45\linewidth]{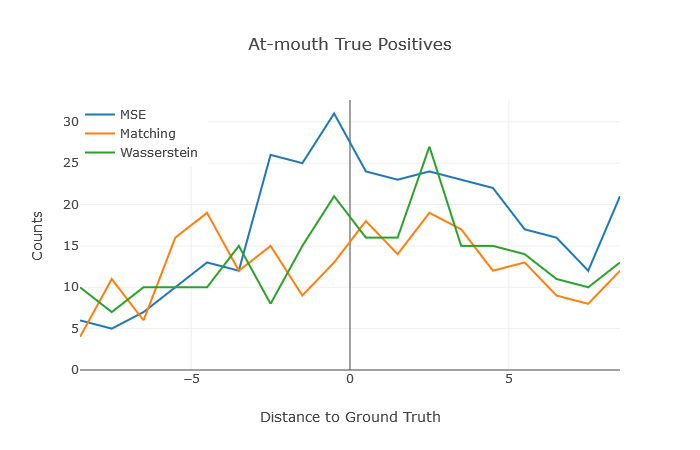} & \includegraphics[width=0.45\linewidth]{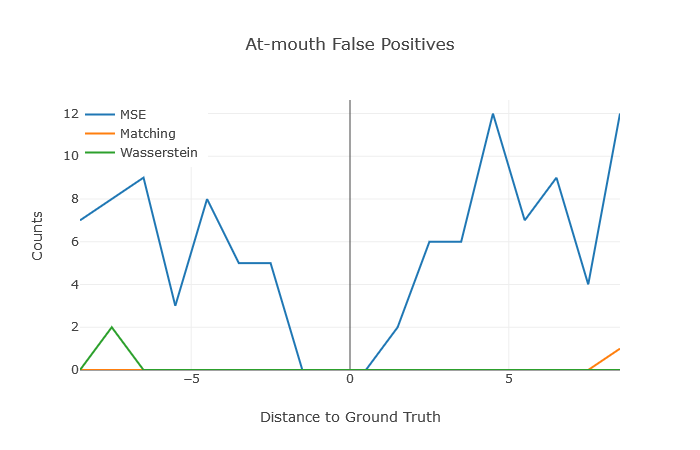} \\
\includegraphics[width=0.45\linewidth]{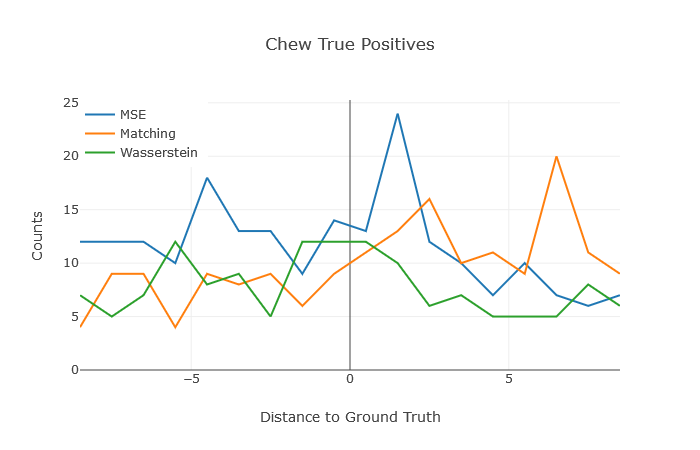} & \includegraphics[width=0.45\linewidth]{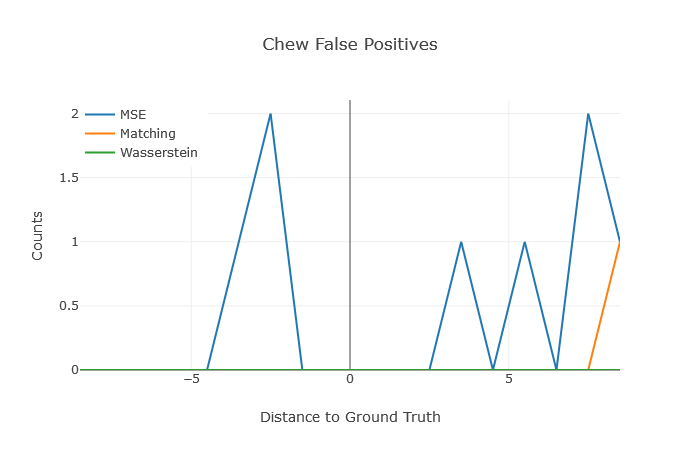} \\
\end{tabular}
\caption{Left column shows the distribution of true positives and the right side the false positives. For these behaviors, the network struggles to predict the start frame accurately.}
\label{fig:fpstps1b}
\end{figure*}

Fig.~\ref{fig:fpstps1a} and Fig.~\ref{fig:fpstps1b} show more examples of the distribution of predictions for each behavior. For most of the behaviors, the number of false positives within $\tau=10$ frames is greatest for the MSE loss.

\begin{figure*}[ht]
\centering
\includegraphics[width=0.8\linewidth]{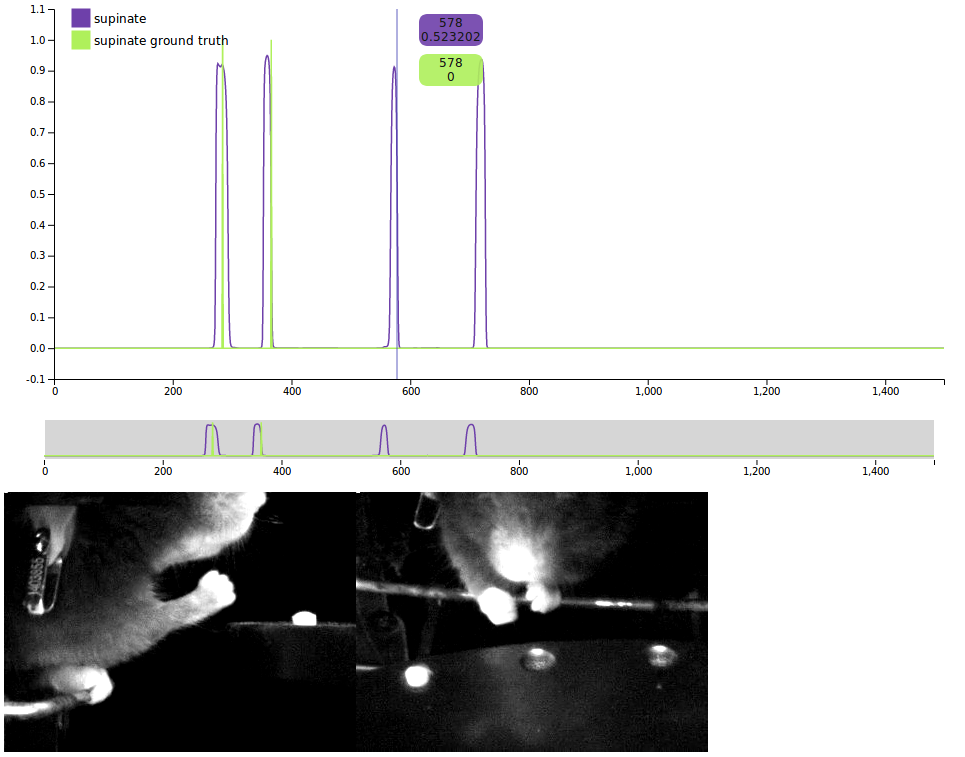}
\caption{An example screen shot of our web based network output viewer for videos. The green line is ground truth and purple is our network's predictions. Here we can mouse over the frames that caused the false positive predictions. The veritical blue line near 600 frames denotes the current visible frame in the video. The purple and green rectangles shows the frame number and scores of that frame. In this case, frame 578 is being viewed and the ground truth supinate score is 0 and the network prediction of the behavior is 0.52. We can see the side of the mouse paw, which is something visible in the mouse supinate behavior, but the paw is quite far from the food pellet.}
\label{fig:viewer_appendix}
\end{figure*}

Fig.~\ref{fig:viewer_appendix} shows a larger screenshot of our visualization tool. The web-based viewer synchronizes the network output score and video frame. The viewer has two main components. A line graph, where the x-axis is video frames and the y-axis is the network output score. The other component is a video viewer, where the frame being shown is the currently selected frame. A frame can be selected by either playing the movie or mousing over the line graph. Being able to slowly, or quickly, mouse over consecutive video frames around curious network outputs helped find software bugs and explore network architectures.

\begin{figure}[ht]
\centering
\begin{tabular}{ll}
\includegraphics[width=0.95\linewidth]{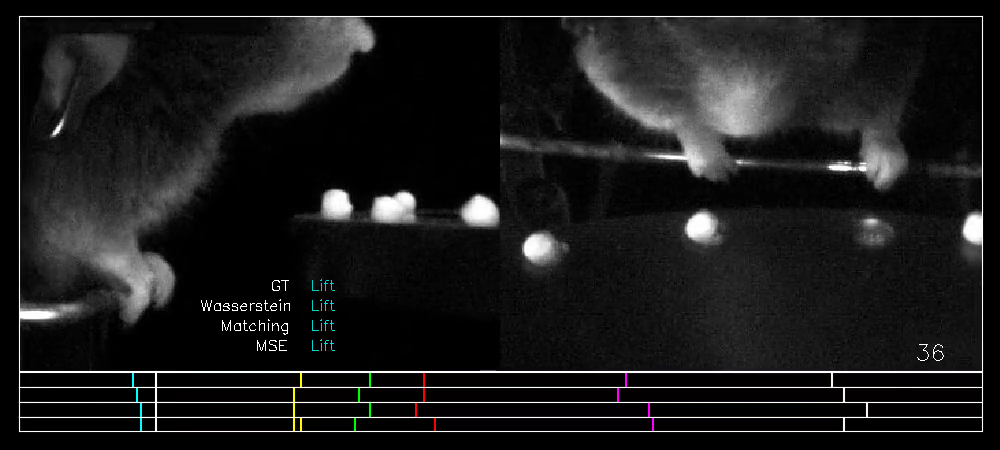}
\end{tabular}
\caption{Example frame from a video showing the classification results. See text for details}
\label{fig:videosnapshot}
\end{figure}

We also provide four videos showing examples of the reach task with the labels predicted by our trained models. Each video shows the front and side views of the mouse with four rows of bars beneath the mouse frames. Figure~\ref{fig:videosnapshot} shows an example frame snapshot. The first row represents the ground truth location of the behaviors, and the following three rows represent the Wasserstein, Matching, and MSE model predictions. Each row has colored vertical bars for each of the behaviors. "Lift" is cyan, "Hand" yellow, "Grab" green, "Supinate" red, "At-mouth" magenta, and "Chew" white. A vertical bar representing the current frame will move across the four rows. Additionally, within the video frame of the mouse reaching task, we add text labels of the predictions. As the video playback frame approaches the frame location of a predicted label, the label name will fade in, and fade out as the playback passes by. The example videos of action start detection are available at \url{http://research.janelia.org/bransonlab/MouseReachData/}. M134\_20150325\_v020.mp4, M174\_20150417\_v031.mp4, and M174\_20150427\_v004.mp4 show the mouse successfully grab and eat the food pellet. In videos M134\_20150325\_v020.mp4 and M174\_20150427\_v004.mp4, all three networks properly predict each behavior, however the MSE loss produces extra false positives. In M174\_20150417\_v031.mp4, all three networks struggle to properly detect each behavior. Both the Matching and MSE losses produce extra false positives, while the Wasserstein loss fails to detect the "Chew" behavior. M134\_20150504\_v018.mp4 shows an example of the mouse failing to grab the food pellet on its first try. The Wasserstein loss properly detects each behavior, while the MSE loss produces a large number of false positives.

